\documentclass[11pt,twocolumn]{article}

\usepackage[T1]{fontenc}
\usepackage{newtxtext}

\usepackage{amsmath,amsthm,amssymb}
\usepackage[nosymbolsc]{newtxmath}
\usepackage{microtype}

\usepackage[margin=0.75in,columnsep=0.3in]{geometry}
\usepackage{titlesec}
\usepackage{enumitem}
\usepackage{needspace}
\usepackage{etoolbox}

\usepackage{booktabs}
\usepackage{array}
\usepackage{colortbl}
\usepackage[labelfont=bf]{caption}

\usepackage{graphicx}
\usepackage{float}
\usepackage{xcolor}
\usepackage{tikz}
\usepackage{pgfplots}
\pgfplotsset{compat=1.18}
\usetikzlibrary{arrows.meta, positioning, shapes.geometric, patterns, calc}

\usepackage{hyperref}
\hypersetup{colorlinks=true, linkcolor=black, citecolor=black, urlcolor=blue}

\definecolor{teacherblue}{RGB}{66, 133, 244}
\definecolor{failred}{RGB}{219, 68, 55}
\definecolor{successgreen}{RGB}{15, 157, 88}
\definecolor{accentgray}{RGB}{128, 128, 128}
\definecolor{sweetspot}{RGB}{200, 230, 200}
\definecolor{varianceblue}{RGB}{100, 149, 237}
\definecolor{biasred}{RGB}{205, 92, 92}

\titleformat{\section}
  {\large\bfseries\color{black}}
  {\thesection}{1em}{}
  [\vspace{-0.5em}\textcolor{accentgray}{\titlerule[0.4pt]}]
\titlespacing*{\section}{0pt}{1.5em}{0.75em}

\preto\section{\needspace{4\baselineskip}}

\newtheorem{definition}{Definition}[section]

\newcommand{\insight}[1]{%
  \begin{center}
    \parbox{0.9\linewidth}{%
      \centering\large\itshape\bfseries #1%
    }
  \end{center}
  \vspace{0.5em}
}

\newcommand{\calloutbox}[1]{%
  \smallskip
  \noindent\fbox{\parbox{0.95\linewidth}{\small #1}}%
  \smallskip
}

\title{\textbf{Hallucinations Live in Variance}}

\author{
Aaron R.\ Flouro \quad Shawn P.\ Chadwick, PhD\\[0.5em]
\normalsize\texttt{research@sparse-tech.com}
}

\date{}

\begin{document}

\maketitle

\begin{abstract}
\noindent Benchmarks measure whether a model is correct. They do not measure whether a model is reliable. This distinction is largely academic for single-shot inference, but becomes critical for agentic AI systems, where a single rephrased prompt can trigger cascading failures in multi-step execution. Yet this form of instability is not captured by existing evaluations.

\emph{Hallucinations live in variance:} they arise when semantically equivalent prompts activate inconsistent internal pathways, producing divergent outputs. Consistent but incorrect outputs reflect bias or missing knowledge; confident guessing reflects calibration failure. Neither constitutes hallucination under this definition. When error is variance-dominated, reducing redundant pathways improves reliability without adding knowledge. We formalize this through Semantic Stability (SS), measured via Paraphrase Consistency (PC@$k$): generate $k$ paraphrases, greedy decode each, compute mode agreement. SS is a diagnostic for variance-driven unreliability, not a method for improving correctness.

We show that a dense Qwen3-0.6B agrees with itself only 23.8\% of the time; at 32\% sparsity, agreement jumps to 55.9\%. A phase diagram reveals the sweet spot where variance reduction outpaces bias accumulation, and regimes where stability collapses onto wrong answers.
\end{abstract}

\noindent\textbf{Index Terms:} Semantic Stability, Knowledge Distillation, Hallucination Analysis, Model Compression, Bias-Variance Decomposition, LLM Evaluation

%==============================================================================
\section{The Missing Axis}
%==============================================================================

Benchmarks answer one question: \emph{Is it correct?} Users experience a different question: \emph{Is it reliable?}

A model that scores 80\% on a reasoning task may still frustrate users who discover that rephrasing their question produces a different answer. The benchmark cannot distinguish a model that reliably gets 80\% right from one that unpredictably scatters across answers. Both score the same. Only one is trustworthy.

This gap matters now more than ever. Interactive deployment means users iterate on prompts in real time. Hallucinations have become an enterprise risk, blocking adoption in high-stakes domains. And compression is no longer optional: edge deployment, latency constraints, and cost pressures make smaller models inevitable.

Terminology. Throughout this paper, \emph{hallucination} refers exclusively to variance-driven instability under semantic perturbation. Outputs that are stable but wrong reflect bias or missing knowledge and are treated as \emph{incorrectness}. Confident guessing in the absence of signal reflects calibration failure. These behaviors are distinct from hallucination and are addressed by correctness or calibration metrics, not Semantic Stability.

This paper is not about improving accuracy. It is about measuring and understanding instability. We introduce a metric that exposes variance-driven unreliability, show how compression can reduce it, and map the regimes where this holds.

%==============================================================================
\section{A Metric for Reliability: Semantic Stability}
%==============================================================================

Semantic Stability is a diagnostic model property: it measures reliability under semantic perturbation and does not itself improve correctness. This is not factual consistency~\cite{kryscinski2020evaluating} (agreement with ground truth) or faithfulness~\cite{maynez2020faithfulness} (agreement with source documents). It is self-agreement under paraphrase. Self-consistency methods~\cite{wang2022selfconsistency} vary sampled reasoning paths to improve accuracy; Semantic Stability varies prompt semantics under deterministic decoding to measure structural disagreement.

\begin{definition}[Paraphrase Consistency and Semantic Stability]
\label{def:ss}
Given a prompt $x$, generate $k$ paraphrases $\{x_1, \ldots, x_k\}$ and decode each greedily. Let $a_i$ denote the response to $x_i$. Paraphrase Consistency is the mode agreement:
\begin{equation}
\mathrm{PC}@k(x) = \frac{\max_a |\{i : a_i = a\}|}{k}
\end{equation}
Semantic Stability is the dataset mean:
\begin{equation}
\mathrm{SS} = \mathbb{E}_x[\mathrm{PC}@k(x)]
\end{equation}
\end{definition}

Greedy decoding~\cite{sutskever2014sequence} is essential. At each step $t$, the model selects the highest-probability token:
\begin{equation}
a_t = \arg\max_{v \in \mathcal{V}} P(v \mid x, a_{<t})
\end{equation}
where $\mathcal{V}$ is the vocabulary and $a_{<t}$ denotes previously generated tokens. This removes sampling noise entirely. Any remaining variance is structural: it lives in the model's internal pathways, not in the random seed. This is generalization variance made visible.

\insight{Hallucinations live in variance. SparseKD reduces variance.}

This holds in variance-dominated regimes only. When bias dominates, when the model confidently commits to wrong answers, compression can make things worse.

Sparse Knowledge Distillation (SparseKD) is a multi-stage compression procedure that combines structured capacity reduction with probability-domain knowledge distillation~\cite{hinton2015distilling}. At each stage, representational degrees of freedom are reduced while the student model is re-anchored to a higher-capacity teacher using log-probabilities rather than hard labels. The purpose of SparseKD is not to add knowledge, but to reduce generalization variance by constraining redundant internal pathways while preserving the dominant predictive signal. The theoretical conditions under which this reduces error are formalized in~\cite{flouro2026sparsekd}. Appendix~A provides a compliance-ready evaluation protocol for computing Semantic Stability on any benchmark.

%==============================================================================
\section{Why Existing Metrics Miss This}
%==============================================================================

\emph{Perplexity} measures how surprised a model is by text. A model can have low perplexity and still produce wildly different outputs for equivalent prompts. Likelihood is not stability.

\emph{Accuracy and Pass@k}~\cite{chen2021evaluating} collapse variance by design. They ask whether \emph{any} of $k$ samples is correct, or whether \emph{the} answer is correct. Neither reveals whether the model consistently produces the same answer. A model that scatters across five different responses, one of which happens to be right, scores identically to one that reliably produces the correct answer.

\emph{Factuality benchmarks} such as TruthfulQA~\cite{lin2021truthfulqa} measure correctness, not self-agreement. A model can be truthful on average while being erratic on individual prompts. These metrics expose \emph{what} the model believes, not \emph{whether} it believes it consistently.

None of these metrics expose variance-driven instability. Semantic Stability does.

%==============================================================================
\section{Related Work and Limitations}
%==============================================================================

Recent work on hallucinations in large language models largely falls into four categories: grounding-based methods, preference-alignment methods, calibration and belief-centric evaluations, and consistency or sampling-based reliability techniques. While each improves correctness or detection under specific assumptions, none directly measure variance-driven instability under semantic perturbation.

\emph{Grounding-based approaches} mitigate hallucinations by augmenting inference with external structure, such as knowledge graphs, tool execution, or code-guided reasoning. Representative examples include code- or graph-mediated reasoning systems that constrain outputs using external symbolic structure or executable programs~\cite{arxiv260104086, arxiv260102739}. These methods improve factual accuracy by restricting information sources, but they operate at inference time, are task-specific, and do not measure whether a model's internal behavior is stable under semantically equivalent prompts.

\emph{Preference alignment and factuality-aware training methods} address hallucinations by modifying training objectives to favor truthful or grounded responses, often using labeled preference data or synthetic hallucinated variants~\cite{arxiv260103027}. While effective for aligning beliefs with reference answers, these approaches primarily optimize correctness and calibration. They do not expose whether a model produces consistent outputs when semantic meaning is held fixed, and therefore cannot distinguish bias-dominated collapse from variance-driven instability.

\emph{Calibration and belief-centric evaluations} frame hallucinations as misbelief or miscalibration problems and assess correctness or confidence relative to ground truth~\cite{arxiv251222416}. These methods assume a single output per prompt and require reference answers. As a result, they cannot observe disagreement across semantically equivalent inputs and do not capture instability that emerges only when prompts are rephrased.

\emph{Consistency and sampling-based methods} attempt to improve reliability by aggregating multiple sampled reasoning paths or responses. Self-consistency techniques in chain-of-thought prompting~\cite{wang2022selfconsistency} and adaptive sampling approaches such as Reliability-Aware Self-Consistency (ReASC)~\cite{arxiv260102970} reduce error by selecting or weighting sampled trajectories. Related work also explores cross-model or cross-response agreement for hallucination detection~\cite{arxiv251112236, arxiv250814314}. These methods rely on stochastic sampling, multiple models, or confidence heuristics, and therefore conflate sampling variance with structural instability.

Recent benchmarking efforts highlight evaluation blind spots. HalluLens introduces a taxonomy of intrinsic and extrinsic hallucinations~\cite{arxiv250417550}, while critiques of lexical hallucination metrics demonstrate the inadequacy of string-level overlap measures~\cite{arxiv250808285}. These works motivate improved evaluation but do not introduce a deterministic measure of self-agreement under semantic perturbation.

Across all of these approaches, a common limitation is that evaluation collapses behavior to a single answer per prompt. Variance is therefore averaged away by design. None of the above methods define or measure self-agreement under semantic perturbation with deterministic decoding.

In contrast, Semantic Stability directly measures variance-driven instability by evaluating agreement across paraphrased prompts under deterministic (greedy) decoding, without relying on ground truth, sampling randomness, or external resources. This exposes failure modes such as stable but incorrect collapse and bias--variance tradeoffs under compression that are invisible to existing benchmarks. The resulting stability--sparsity phase diagram provides a unifying lens for understanding when compression improves reliability and when it degrades it.

%==============================================================================
\section{Variance as the Mechanism}
%==============================================================================

Dimensionality reduction as a tool for stability is well understood in other fields. In statistics and signal processing, Principal Component Analysis (PCA)~\cite{hotelling1933pca} and its computational counterpart Singular Value Decomposition (SVD)~\cite{golub2013matrix} decompose a system into orthogonal components ordered by energy or variance. Retaining the dominant components preserves signal, while discarding low-energy components suppresses noise and instability. In finance, PCA is used to remove unstable factors from portfolios to reduce sensitivity to small perturbations. In engineering, SVD is used to eliminate high-frequency modes that introduce jitter without contributing meaningful structure.

The effect of these methods is not improved information content, but improved reliability. Systems with fewer unstable modes respond more consistently to small changes in input.

SparseKD plays an analogous role for language models. Large neural networks contain many overlapping, redundant internal pathways that can represent similar concepts. When a prompt is paraphrased, slightly different pathways may activate, producing different outputs even when the semantic meaning is unchanged. By reducing representational degrees of freedom and re-anchoring the remaining structure through knowledge distillation, SparseKD suppresses these unstable pathways. When error is variance-dominated, this reveals a tighter response distribution. When error is bias-dominated, the same process can collapse the model onto an incorrect answer.

This analogy is not exact, but the principle is the same: removing redundant degrees of freedom reduces variance, not knowledge.

This is the bias-variance decomposition~\cite{geman1992biasvariance} applied to neural networks:
\begin{equation}
\text{Error} = \underbrace{\text{Bias}^2}_{\text{deviation from truth}} + \underbrace{\text{Variance}}_{\text{internal jitter}}
\end{equation}

The formal framework in \emph{Sparse Knowledge Distillation}~\cite{flouro2026sparsekd} proves the condition under which a student outperforms its teacher:
\begin{equation}
\text{Bias}^2_{\text{student}} + \text{Var}_{\text{student}} < \text{Var}_{\text{teacher}}
\end{equation}

This paper shows what that condition \emph{looks like}. When compression shrinks the hypothesis space, it can reduce variance, or it can introduce bias. The outcome depends on what you remove and how you remove it. Unlike most compression techniques, which accept increased bias or likelihood degradation, SparseKD can reduce variance sufficiently to improve perplexity when the $\Delta\mathrm{Var} > \Delta\mathrm{Bias}^2$ condition holds.

\begin{figure}[H]
\centering
\begin{tikzpicture}[scale=0.75]
\begin{axis}[
    name=panel_a,
    width=0.30\textwidth,
    height=3.2cm,
    xmin=-3, xmax=3,
    ymin=0, ymax=0.5,
    axis lines=none,
    title={\small\textbf{(A) Teacher}},
    title style={yshift=-0.5em},
    clip=false
]
\addplot[domain=-3:3, samples=100, thick, color=teacherblue, fill=teacherblue, fill opacity=0.2] {0.4*exp(-x^2/2)};
\draw[teacherblue, dashed, thick] (axis cs:0,0) -- (axis cs:0,0.42);
\node[teacherblue, font=\footnotesize] at (axis cs:0,0.47) {$\mu$};
\draw[<->, teacherblue, thick] (axis cs:-1,0.15) -- (axis cs:1,0.15);
\node[teacherblue, font=\footnotesize] at (axis cs:0,0.08) {$\sigma$};
\end{axis}

\begin{axis}[
    name=panel_b,
    at={($(panel_a.east)+(0.4cm,0)$)},
    anchor=west,
    width=0.30\textwidth,
    height=3.2cm,
    xmin=-3, xmax=3,
    ymin=0, ymax=0.5,
    axis lines=none,
    title={\small\textbf{(B) One-Shot}},
    title style={yshift=-0.5em},
    clip=false
]
\addplot[domain=-3:3, samples=100, thick, color=failred, fill=failred, fill opacity=0.2] {0.25*exp(-(x-0.8)^2/1.5)};
\draw[failred, dashed, thick] (axis cs:0.8,0) -- (axis cs:0.8,0.27);
\node[failred, font=\footnotesize] at (axis cs:0.8,0.32) {$\mu'$};
\draw[accentgray, dotted] (axis cs:0,0) -- (axis cs:0,0.42);
\node[accentgray, font=\tiny] at (axis cs:-0.6,0.35) {shifted};
\end{axis}

\begin{axis}[
    name=panel_c,
    at={($(panel_b.east)+(0.4cm,0)$)},
    anchor=west,
    width=0.30\textwidth,
    height=3.2cm,
    xmin=-3, xmax=3,
    ymin=0, ymax=0.5,
    axis lines=none,
    title={\small\textbf{(C) Staged}},
    title style={yshift=-0.5em},
    clip=false
]
\addplot[domain=-3:3, samples=100, thick, color=successgreen, fill=successgreen, fill opacity=0.2] {0.45*exp(-x^2/0.8)};
\draw[successgreen, dashed, thick] (axis cs:0,0) -- (axis cs:0,0.47);
\node[successgreen, font=\footnotesize] at (axis cs:0,0.52) {$\mu$};
\draw[<->, successgreen, thick] (axis cs:-0.6,0.2) -- (axis cs:0.6,0.2);
\node[successgreen, font=\footnotesize] at (axis cs:0,0.12) {$\sigma'$};
\end{axis}
\end{tikzpicture}
\caption{Compression as variance reduction. \textcolor{teacherblue}{(A)}~Teacher has high variance around mean $\mu$. \textcolor{failred}{(B)}~One-shot pruning shifts the mean and distorts the distribution. \textcolor{successgreen}{(C)}~Staged pruning preserves $\mu$ while reducing variance.}
\label{fig:variance_reduction}
\end{figure}
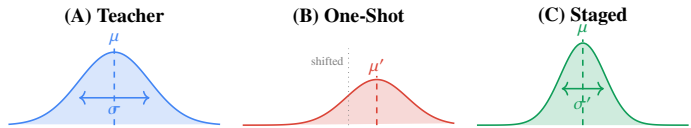

%==============================================================================
\section{The Phase Diagram: Stability vs Sparsity}
%==============================================================================

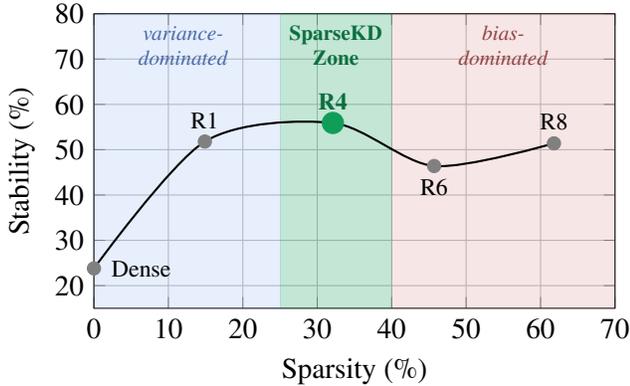
\begin{figure}[t]
\centering
\begin{tikzpicture}
\begin{axis}[
    width=\columnwidth,
    height=5.5cm,
    xlabel={Sparsity (\%)},
    ylabel={Stability (\%)},
    xmin=0, xmax=70,
    ymin=15, ymax=80,
    grid=both,
    grid style={line width=0.2pt, draw=gray!30},
    major grid style={line width=0.4pt, draw=gray!50},
    xtick={0,10,20,30,40,50,60,70},
    ytick={20,30,40,50,60,70,80},
    legend pos=south east,
    legend style={font=\footnotesize, draw=none, fill=white, fill opacity=0.8},
    clip=false
]
% Variance-dominated region (blue)
\addplot[fill=varianceblue, fill opacity=0.15, draw=none, forget plot] coordinates {(0,15) (0,80) (25,80) (25,15)} \closedcycle;
% SparseKD Zone (green)
\addplot[fill=successgreen, fill opacity=0.25, draw=none, forget plot] coordinates {(25,15) (25,80) (40,80) (40,15)} \closedcycle;
% Bias-dominated region (red)
\addplot[fill=biasred, fill opacity=0.15, draw=none, forget plot] coordinates {(40,15) (40,80) (70,80) (70,15)} \closedcycle;
% Data curve
\addplot[thick, color=black, smooth, tension=0.5] coordinates {(0, 23.8) (14.9, 51.8) (32.1, 55.9) (45.7, 46.4) (61.8, 51.4)};
\addplot[only marks, mark=*, mark size=2.5pt, color=accentgray] coordinates {(0, 23.8) (14.9, 51.8) (45.7, 46.4) (61.8, 51.4)};
\addplot[only marks, mark=*, mark size=4pt, color=successgreen] coordinates {(32.1, 55.9)};
% Labels
\node[font=\footnotesize, anchor=west] at (axis cs:1, 23.8) {Dense};
\node[font=\footnotesize, anchor=south] at (axis cs:14.9, 52.5) {R1};
\node[font=\footnotesize\bfseries, anchor=south, color=successgreen!70!black] at (axis cs:32.1, 56.6) {R4};
\node[font=\footnotesize, anchor=north] at (axis cs:45.7, 45.7) {R6};
\node[font=\footnotesize, anchor=south] at (axis cs:61.8, 52.1) {R8};
% Region labels (positioned at top to avoid overlap with data points)
\node[font=\scriptsize, color=varianceblue!70!black, align=center] at (axis cs:12, 73) {\textit{variance-}\\\textit{dominated}};
\node[font=\scriptsize\bfseries, color=successgreen!70!black, align=center] at (axis cs:32.5, 73) {SparseKD\\Zone};
\node[font=\scriptsize, color=biasred!70!black, align=center] at (axis cs:55, 73) {\textit{bias-}\\\textit{dominated}};
\end{axis}
\end{tikzpicture}
\caption{The SparseKD stability phase diagram. Stability peaks at 32\% sparsity (R4). \textcolor{varianceblue}{Blue}: variance-dominated regime where compression tightens distributions. \textcolor{successgreen}{Green}: optimal SparseKD zone. \textcolor{biasred}{Red}: bias-dominated regime where over-pruning collapses onto wrong answers.}
\label{fig:phase_diagram}
\end{figure}

Theory predicts a stability peak at moderate sparsity. Semantic Stability measures variance, not correctness, and must always be reported alongside accuracy. We tested this on Qwen3-0.6B using multi-stage pruning. The x-axis is sparsity (percentage of parameters removed). The y-axis is Semantic Stability (PC@$k$). Figure~\ref{fig:phase_diagram} shows the results.

The curve is non-monotone. The dense model gives the same answer only 23.8\% of the time. At 32\% sparsity (R4), stability jumps to 55.9\%, a 32-point gain from \emph{removing} parameters. Beyond 45\% sparsity, stability declines.

\begin{table}[t]
\centering
\small
\setlength{\tabcolsep}{4pt}
\begin{tabular}{@{}lcccc@{}}
\toprule
\textbf{Stage} & \textbf{Sparsity} & \textbf{Stability} & \textbf{$\boldsymbol{\Delta}$} & \textbf{PPL} \\
\midrule
Dense & 0.0\% & 23.8\% & -- & 35.87 \\
R1 & 14.9\% & 51.8\% & +28.0 & 28.90 \\
\rowcolor{sweetspot!50}
\textbf{R4} & \textbf{32.1\%} & \textbf{55.9\%} & \textbf{+32.1} & \textbf{28.37} \\
R6 & 45.7\% & 46.4\% & +22.6 & 29.99 \\
R8 & 61.8\% & 51.4\% & +27.6 & 32.90 \\
\bottomrule
\end{tabular}
\caption{Qwen3-0.6B stability sweep. R4 achieves peak stability with minimal perplexity cost. $\Delta$ shows improvement over dense baseline.}
\label{tab:sweep}
\end{table}

The interpretation maps directly onto bias-variance. In the variance-dominated region (left side), compression removes redundant pathways and tightens the response distribution. In the bias-dominated region (right side), aggressive capacity reduction without sufficient distillation coverage fails to preserve critical pathways, and the model collapses onto wrong answers.

R4 sits in the sweet spot: variance reduction outpaces bias accumulation. Notably, at the stability peak (R4), perplexity improves relative to the dense baseline, indicating a net tightening of the predictive distribution rather than the typical accuracy-efficiency tradeoff observed in compression.

%==============================================================================
\section{Adding SS to Any Benchmark}
%==============================================================================

Semantic Stability is benchmark-agnostic. It requires no new labels, no new datasets, no architectural changes. The procedure is simple:

\begin{enumerate}[nosep,leftmargin=1.5em]
    \item Paraphrase: Generate $k$ semantically equivalent variants of each prompt.
    \item Decode: Greedy decode each variant (no sampling noise).
    \item Compute: Calculate mode agreement across responses.
\end{enumerate}

This can be layered onto GSM8k, TruthfulQA, MMLU~\cite{hendrycks2020mmlu}, or any internal evaluation. The result is a second axis: not just ``how often is the model correct?'' but ``how often does the model agree with itself?''

This turns hallucination risk into a measurable quantity. Instead of hoping a model is reliable, you can measure it. Instead of discovering instability in production, you can expose it in evaluation.

While Semantic Stability exposes variance-driven instability, reaching the stability peak requires structured variance reduction. In practice, this is achieved through the multi-stage probability-domain temperature scaling and compression protocols defined in \emph{Sparse Knowledge Distillation}~\cite{flouro2026sparsekd}.

%==============================================================================
\section{Stable Does Not Mean Correct}
\label{sec:stable_not_correct}
%==============================================================================

This is one of the most important findings. Stability and correctness are orthogonal axes.

\begin{table}[t]
\centering
\small
\setlength{\tabcolsep}{3pt}
\begin{tabular}{@{}llccc@{}}
\toprule
\textbf{Model} & \textbf{Probe} & \textbf{SS} & \textbf{\%PC$<$0.5} & \textbf{\%PC$\geq$0.8} \\
\midrule
Dense & GSM8k & 0.296 & 92\% & 0\% \\
Dense & Facts & 0.440 & 64\% & 10\% \\
\rowcolor{sweetspot!50}
\textbf{R4} & \textbf{GSM8k} & \textbf{0.316} & \textbf{82\%} & 0\% \\
R4 & Facts & 0.232 & 90\% & 0\% \\
\bottomrule
\end{tabular}
\caption{Semantic Stability on Qwen3-0.6B. R4 improves SS on GSM8k but degrades it on Facts. $k{=}10$ paraphrases, $N{=}50$ prompts per probe.}
\label{tab:ss}
\end{table}

On GSM8k~\cite{cobbe2021gsm8k} (arithmetic reasoning), R4 improves SS by +0.02 over Dense. The dense model scatters across answer formats; compression tightens the distribution.

On Facts (factual recall), R4 exhibits lower Semantic Stability ($-0.21$). The dense model already contains the relevant factual knowledge, but the distillation signal used during compression was insufficient to consistently re-anchor all factual pathways. As a result, variance reduction in this regime collapses the model onto incomplete or incorrect factual retrieval routes, yielding stable but wrong answers. Increasing coverage of the distillation corpus or strengthening teacher supervision would be expected to preserve factual stability under the same compression regime.

A concrete example makes this vivid:

\calloutbox{%
\emph{Q:} ``What is the largest planet?''\\[2pt]
\emph{Dense} (PC=0.20): Jupiter, Neptune, Saturn, \ldots\\
\emph{R4} (PC=0.60): ``the largest planet is the sun'' $\times$6
}

R4 is more stable. R4 is also wrong. The model collapsed onto a confident, incorrect answer. This is bias, not variance.

The lesson: Semantic Stability captures variance-driven instability, not correctness. A model can become more stable by collapsing onto a wrong answer. Never optimize SS alone; always pair with accuracy.

%==============================================================================
\section{Implications}
%==============================================================================

\emph{Prompting is noise control.} Users learn to phrase questions carefully because models behave inconsistently. Variance reduction addresses this at the source: prompts that are semantically equivalent begin to behave equivalently.

\emph{Compression is a reliability tool.} In variance-dominated regimes, smaller models are not just cheaper. They are more consistent. This reframes the cost-quality tradeoff.

\emph{Semantic Stability is an evaluation layer.} Adding SS to any benchmark exposes a failure mode that accuracy alone cannot capture. It makes reliability visible.

In enterprise and regulated settings, reliability is no longer a secondary concern. The EU AI Act (2024) and NIST AI 600-1 guidelines now require high-risk AI systems to exhibit predictable behavior and provide transparent evidence of failure modes. Semantic Stability provides a measurable reliability signal that can be audited alongside accuracy, exposing variance-driven instability before deployment. This aligns evaluation with the requirements of agentic workflows, where consistency is often more critical than average correctness.

%==============================================================================
\section{Conclusion}
%==============================================================================

Benchmarks measure correctness. Users experience reliability. This paper introduces the missing axis.

What you can do today:
\begin{itemize}[nosep,leftmargin=1.5em]
    \item Add Semantic Stability to any evaluation by paraphrasing prompts and measuring mode agreement
    \item Use the phase diagram to find compression sweet spots for your models
    \item Pair SS with accuracy: stability without correctness is bias collapse
\end{itemize}

Instability is not mysterious. It is variance. And variance, unlike knowledge, can be removed. The math lives in \emph{Sparse Knowledge Distillation}~\cite{flouro2026sparsekd}. The intuition lives here.

\emph{Hallucinations live in variance. SparseKD reduces variance.}

%==============================================================================
\section*{Acknowledgments}
%==============================================================================

The authors gratefully acknowledge the collaborative environment at SparseTech that made this research possible. The theoretical and computational developments presented in this paper are part of an ongoing SparseTech research initiative on sparse knowledge distillation for large language models. Patent Pending.

\bibliographystyle{plain}
\bibliography{paper1.5_viral_companion}

\onecolumn
\appendix
\renewcommand{\thesection}{}
\section*{Appendix A: Semantic Stability Evaluation Protocol}

\subsection*{Purpose}

Quantify AI system robustness under semantic perturbation to support risk management documentation, conformity assessment, and deployment decisions per EU AI Act Article 15 (Accuracy, Robustness and Cybersecurity) and NIST AI RMF MEASURE function (2.5 Valid and Reliable, 2.6 Robustness).

\subsection*{Scope}

This protocol addresses the robustness characteristic of trustworthy AI by measuring output consistency under equivalent input conditions. It is applicable across the AI system lifecycle for:

\begin{itemize}[nosep]
    \item Pre-deployment conformity assessment
    \item Continuous post-market monitoring
    \item Technical documentation requirements
    \item Third-party audit support
\end{itemize}

\subsection*{Requirements}

\begin{itemize}[nosep]
    \item AI system under evaluation
    \item Paraphrase generator (LLM or template-based)
    \item Deterministic inference (temperature = 0)
    \item Evaluation dataset (existing benchmark or domain-specific)
\end{itemize}

\subsection*{Procedure}

\begin{enumerate}[nosep]
    \item \textbf{Select} $N$ prompts from evaluation set (recommend $N \geq 100$)
    \item \textbf{Generate} $k$ semantically equivalent input variants per prompt (recommend $k = 10$)
    \item \textbf{Execute} deterministic inference on all $k$ variants
    \item \textbf{Compute} PC@$k$ per prompt using Equation 1
    \item \textbf{Document} dataset mean (SS), plus \%PC$<$0.5 and \%PC$\geq$0.8
\end{enumerate}

\subsection*{Reporting Template}

\begin{center}
\begin{tabular}{@{}lcccccc@{}}
\toprule
\textbf{AI System} & \textbf{Evaluation Domain} & \textbf{N} & \textbf{k} & \textbf{SS} & \textbf{\%PC$<$0.5} & \textbf{\%PC$\geq$0.8} \\
\midrule
 & & & & & & \\
\bottomrule
\end{tabular}
\end{center}

\subsection*{Risk Classification Guidance}

\begin{center}
\begin{tabular}{@{}lll@{}}
\toprule
\textbf{SS Range} & \textbf{Robustness Level} & \textbf{Risk Management Recommendation} \\
\midrule
$< 0.3$* & Insufficient & Not suitable for deployment; requires mitigation \\
$0.3$--$0.6$* & Limited & Acceptable for limited-risk applications with monitoring controls \\
$0.6$--$0.9$* & Substantial & Acceptable for high-risk applications with human oversight \\
$> 0.9$* & High & Minimum threshold for safety components in high-risk AI systems \\
\bottomrule
\end{tabular}
\end{center}

\vspace{0.5em}
\noindent\textit{* Semantic Stability measures robustness (output consistency), not accuracy (output correctness). Per NIST AI RMF MEASURE 2.5, validity and reliability assessments require both metrics. SS must be paired with domain-specific accuracy validation. An AI system can achieve high SS by consistently producing incorrect outputs. Do not optimize SS in isolation.}

\subsection*{Formal Definitions}

\textbf{Paraphrase Consistency (PC@$k$):}
\begin{equation*}
\mathrm{PC}@k(x) = \frac{\max_a |\{i : a_i = a\}|}{k}
\end{equation*}

\textbf{Semantic Stability (SS):}
\begin{equation*}
\mathrm{SS} = \mathbb{E}_x[\mathrm{PC}@k(x)]
\end{equation*}

\subsection*{Regulatory Alignment}

\begin{center}
\begin{tabular}{@{}lll@{}}
\toprule
\textbf{Framework} & \textbf{Relevant Provision} & \textbf{SS Protocol Contribution} \\
\midrule
EU AI Act & Article 15(1) -- Robustness & Quantifies robustness to input perturbation \\
EU AI Act & Article 15(3) -- Technical solutions & Measurable mitigation for consistency failures \\
NIST AI RMF & MEASURE 2.5 -- Valid and Reliable & Reliability metric for output consistency \\
NIST AI RMF & MEASURE 2.6 -- Robustness & Perturbation testing methodology \\
ISO/IEC 42001 & 6.1.4 -- AI risk assessment & Reproducible risk quantification \\
\bottomrule
\end{tabular}
\end{center}

\subsection*{Citation}

When referencing this protocol in technical documentation or conformity assessment:

\begin{quote}
Flouro, A. R. \& Chadwick, S. P. (2026). Hallucinations Live in Variance. \textit{arXiv preprint}.
\end{quote}

\end{document}